\documentclass[conference, a4paper]{IEEEtran}
\IEEEoverridecommandlockouts
\usepackage{cite}
\usepackage{amsmath,amssymb,amsfonts}
\usepackage{graphicx}
\graphicspath{{figures/}}
\usepackage{textcomp}
\usepackage{xcolor}
\usepackage[outdir=./]{epstopdf}

\usepackage{colortbl}
\usepackage{algorithm} 
\usepackage{algpseudocode}

\def\BibTeX{{\rm B\kern-.05em{\sc i\kern-.025em b}\kern-.08em
    T\kern-.1667em\lower.7ex\hbox{E}\kern-.125emX}}
\begin{document}

\title{Continual Unsupervised Domain Adaptation for Semantic Segmentation using a Class-Specific Transfer\\
\thanks{This publication was created as part of the research project "KI Delta Learning" (project number: 19A19013R) funded by the Federal Ministry for Economic Affairs and Energy (BMWi) on the basis of a decision by the German Bundestag.}
}

\author{\IEEEauthorblockN{Robert A. Marsden, Felix Wiewel, Mario D\"obler, Yang Yang and Bin Yang}
	\IEEEauthorblockA{\textit{Institute of Signal Processing and System Theory,}
		\textit{University of Stuttgart}\\
		Stuttgart, Germany \\
		Email:    \{robert.marsden, felix.wiewel, mario.doebler, bin.yang\}@iss.uni-stuttgart.de}}


\maketitle


\begin{abstract}
In recent years, there has been tremendous progress in the field of semantic image segmentation. However, one remaining challenging problem is that segmentation models do not generalize to unseen domains. To overcome this problem, one either has to label lots of data covering the whole variety of possible domains, which is often infeasible in practice, or apply unsupervised domain adaptation (UDA), only requiring labeled source data. In this work, we focus on UDA and additionally address the case of adapting not only to a single domain, but to a sequence of target domains. This requires mechanisms preventing the model from forgetting its previously learned knowledge.

To adapt a segmentation model to a target domain, we follow the idea of utilizing light-weight style transfer to convert the style of labeled source images into the style of the target domain, while retaining the source content. To mitigate the distributional shift between the source and the target domain, the model is fine-tuned on the transferred source images in a second step. Existing light-weight style transfer approaches relying on adaptive instance normalization (AdaIN) or Fourier transformation (FDA) still lack performance and do not substantially improve upon common data augmentation, such as color jittering. The reason for this is that these methods do not focus on region- or class-specific differences, but mainly capture the most salient style. Therefore, we propose a simple and light-weight framework that incorporates two class-conditional AdaIN layers. To extract the class-specific target moments needed for the transfer layers, we use unfiltered pseudo-labels, which we show to be an effective approximation compared to real labels. We extensively validate our approach (CACE) on a synthetic sequence and further propose a challenging sequence consisting of real domains. CACE outperforms existing methods visually and quantitatively.
\end{abstract}



\section{Introduction}
Deep convolutional neural networks have significantly contributed to the large success in semantic image segmentation \cite{Deeplabv2, DeepLabv3+, HRNet}. Since this task assigns a class to each pixel in an image, it is well suited for complex applications like automated driving, which require a detailed scene analysis. However, as even state-of-the-art segmentation architectures usually experience a drastic performance drop when they are exposed to data originating from a different distribution than during training, perception in a constantly changing environment such as the real world is still a major challenge. An example for a constant change is the illumination of a scene, but there can also be changes in the current weather situation, where the surroundings can be completely covered in snow or fog within a few minutes. Although the performance degradation can be avoided by collecting and manually labeling new data from the unseen domains, this involves a tremendous amount of data collection and labeling. For example, in \cite{Cityscapes} it is reported that segmenting one single image took around 90 minutes on average. To circumvent manual labeling and still adapt the model to new domains, unsupervised domain adaptation (UDA) can be used. UDA methods attempt to transfer knowledge from a labeled source domain to an unlabeled target domain. This is achieved by mitigating the discrepancy between the source and target distribution in the input space \cite{CYCADA, Bidirectional, FDA, ACE}, the feature space \cite{FCNs, ActivationMatching, CAG, CLST}, the output space \cite{AdaptSegNet, ADVENT, PatchAlign, CLAN}, or even in several spaces in parallel. Nevertheless, the goal of UDA is not sufficient because it neglects the performance on the source domain and focuses only on the target results. Consequently, UDA is particularly useful for the well-studied synthetic-to-real scenario \cite{AdaptSegNet, DACS, CLST}, where the perception system never encounters the synthetic domain again. In the context of automated driving, however, the environment to which the model has just been adapted may return to a previously seen target domain or the source domain. This would not be a problem if neural networks did not suffer from the phenomenon of catastrophic forgetting (CF) \cite{CF}. CF occurs when a model is sequentially trained on a series of domains and is characterized by a decrease in performance on the previous domains while being trained on the current target domain. Although CF could be avoided by saving some target images in a memory (rehearsal), the storage size can be limited during deployment in an automated vehicle. Therefore, an adaptation method should not only be memory efficient, but also prevent forgetting to increase the adaptation efficiency. 

\begin{figure*}[t]
\begin{center}
\def\svgwidth{510pt}
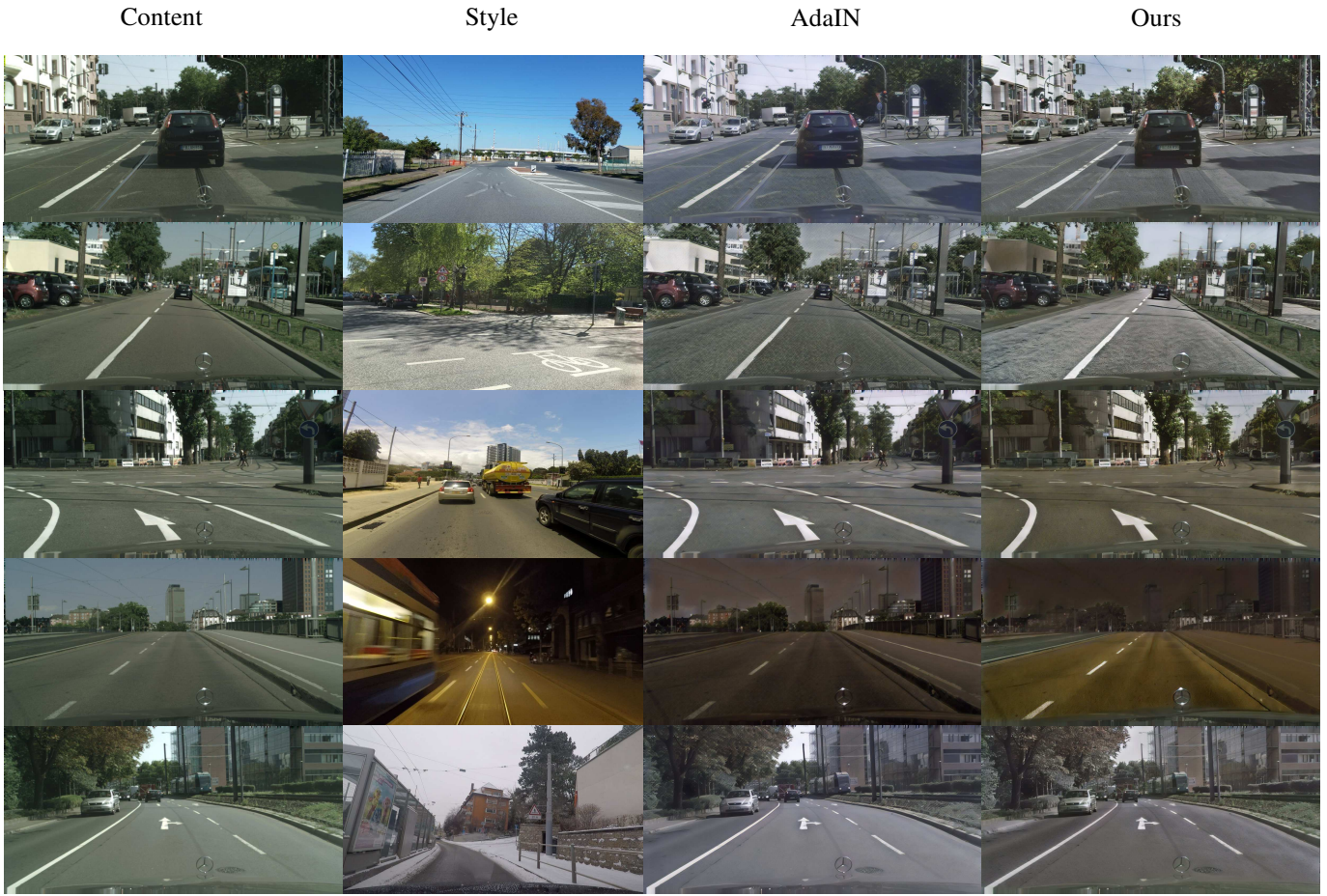
\end{center}
\caption{From the left column to the right column: Content images, style images, stylized images generated with the style transfer network from the ACE framework (AdaIN), stylized images generated with our network.}
\label{fig:image_comparison}
\end{figure*}

Having these requirements in mind, we consider the setting of continual UDA, i.e., we adapt a model pre-trained in a labeled source domain to a sequence of unlabeled target domains while preventing forgetting. Following \cite{ACE}, the sequence contains domains with varying illumination, seasonal, and weather conditions similar to those that could be encountered in the real world. In contrast to \cite{ACE}, we do not only consider a sequence with purely synthetic domains, but also introduce a more sophisticated scenario consisting of only real-world domains. To adapt the model to the current target domain while preventing it from forgetting its previously learned knowledge, we build our work upon the \textit{Adapting to Changing Environments} (ACE) framework \cite{ACE}. A key component of ACE is a light-weight style transfer (ST) network that uses adaptive instance normalization (AdaIN) \cite{AdaIN} to transfer labeled source images into the style of the current target domain. The style transfer is achieved by first renormalizing a source feature map to have the same channel-wise mean and standard deviation as a target feature map before it is subsequently decoded again \cite{AdaIN}. Now, by training with transferred source images, the approach reduces the distribution shift between the source and target domain. However, methods that rely on light-weight style transfer have some major drawbacks that limit their capability to outperform common augmentations, such as color jittering, in a realistic evaluation setting. The most noticeable drawback of ACE, for example, arises directly from its use of the AdaIN layer. Since the renormalization in the style transfer network is applied  to the whole feature map at once (global) rather than to areas of a specific class (class-specific), the newly created image captures only the most dominant aspect of the target style, while missing the exact class-specific differences. This can also be seen in the third column of Fig. \ref{fig:image_comparison} that shows a few transferred source images. Clearly, ACE's AdaIN model does not take into account the individual class modes, since it colors the road bluish, for example, or simply darkens the image. Furthermore, the style transfer network of ACE introduces artifacts, which may again negatively influence the domain alignment. This becomes evident when looking at the license plate of the black car in the first row of Fig. \ref{fig:image_comparison}, which now shows distorted text.

Therefore, we now introduce CACE. CACE overcomes the aforementioned drawbacks by conditioning the style transfer on each class. Further, we slightly modify the style transfer network by inserting a skip connection into the network with a second class-conditional AdaIN layer working on high resolution feature maps. This helps to overcome the artifacts and enables a more precise class-wise style transfer, especially for small objects. 

We summarize our contributions as follows: 
\begin{itemize}
\item We introduce CACE, a simple light-weight framework that relies on a class-specific style transfer. It consistently outperforms comparable methods when incrementally learning new target domains in an unsupervised manner. 
\item We propose a memory efficient variant of our method for memory-restricted applications. 
\item We empirically show that using color jittering as data augmentation performs roughly as good as several more sophisticated methods when adapting a model to realistic environmental changes. 
\item We validate our approach not only for one completely synthetic sequence, but also introduce a more challenging sequence that consists of purely real domains.  
\end{itemize}

\section{Related works}

\subsection{Image-to-Image Translation} 
Although generative adversarial networks (GANs) \cite{GAN} were originally introduced for synthesizing new images, they can also be used for image-to-image translation, where an image is altered only in its style while preserving the content. Typically, the style transfer is accomplished by using a cycle-consistency loss \cite{CycleGAN}, which encourages the network to preserve the content during stylization. However, other approaches use either the shared latent space assumption \cite{UNIT, MUNIT} or rely on contrastive learning \cite{CUT}.
Although the transferred images of GANs can look appealing, they require a significant amount of computational power and cannot be easily extended to our continual, memory- and adaptation-efficient setting.  In contrast, neural style transfer methods are well suited for our environment, as they are generally less computationally intensive and easy to train. \cite{Gram, PerceptualLoss}, for example, simply align the feature statistic of Gram matrices at different network depths. \cite{AdaIN, ST_MMD} match the first and second order moments of content and a style feature maps. Further, \cite{ST_MMD} and \cite{ST_CMD} interpret the neural style transfer as a domain adaptation problem and minimizes the maximum mean discrepancy \cite{MMD} and the central moment discrepancy \cite{CMD}, respectively.

\subsection{Unsupervised Domain Adaptation} 
\subsubsection{Adversarial Learning}
If the appearance of two domains differs mainly in texture and color and not in geometric structure, it is also possible to leverage image-to-image translation for domain adaptation. Some approaches use a CycleGan-based architecture \cite{CycleGAN} to transform source images to the target style \cite{CYCADA, Bidirectional}. Since  the transformation is content-preserving, the source labels can be reused to adapt the segmentation model with target-like images. To avoid the computationally demanding CycleGAN framework, other approaches like \cite{ACE} rely on adaptive instance normalization \cite{AdaIN} to transfer the source images. 

Another line of work uses adversarial training to align feature distributions \cite{FCNs, CYCADA, Proxy, NoMoreDiscrimination}. In this case, the discriminator becomes a domain classifier that predicts to which domain a feature belongs. \cite{AdaptSegNet, ADVENT, PatchAlign} extend this idea and perform distributional alignment in the output space at the pixel and patch level, respectively. This adaptation strategy is also commonly used either as a basic component or for warm-up \cite{SIM, MRNet, CAG}.

\subsubsection{Self-Training}
A very successful strategy, also found in several state-of-the-arts methods for UDA, is self-training (ST) \cite{IAST, DACS, CLST, FDA, SIM}. It basically converts the target predictions into pseudo-labels to minimize the cross-entropy. However, ST only works if high quality pseudo-labels (PL) can be provided. One line of work attempts to improve their quality by using an ensemble of three models \cite{FDA}, a memory-efficient temporal ensemble \cite{CLST}, or an average of different outputs of the same network \cite{MRNet}. Nevertheless, most ST-based approaches try to filter out the noisy target predictions by using reliability measures. \cite{IAST, SIM, CBST, CCM, MaxSquares}, for example, use the softmax output as confidence measure, assuming that a higher prediction probability is coupled with higher accuracy. \cite{RectifyingPL} explicitly estimates the pixel-wise uncertainty of the predictions. \cite{Proxy} relies on the confidence of a segmentation model in combination with two discriminators. \cite{CAG} uses the distance in the feature space, which they found to be less biased towards the source domain compared to the final segmentation head. 

\subsection{Continual Learning}
Continual learning addresses the problem of incrementally learning new concepts while preventing catastrophic forgetting \cite{CF}. It has been intensively studied for the problem of classification, where the literature can be broadly divided into regularization-based \cite{EWC, SI, RI}, memory-based \cite{GenReplay, Remind}, and model-based \cite{PNN, DEN} approaches. However, for the task of semantic image segmentation and the setting of incrementally learning a sequence of new environments in an unsupervised manner, the literature is still very sparse. \cite{ETM} uses a double hinge adversarial loss in combination with a small target specific memory to further prevent CF. \cite{ACE} relies on light-weight style transfer to create target-like source images used for the adaptation. In addition, the approach mitigates CF by first saving and then replaying previous target styles during training.


\section{Method}
\subsection{Definitions} 
The goal of our approach is to incrementally learn a sequence of domains $\{\mathcal{D}_t\}_{t=0}^T$, without forgetting previously acquired knowledge. The sequence starts with the source domain $\mathcal{D}_0$, whose $N^0$ images and segmentation maps $X^0 = \{\mathbf{x}_i^0, \mathbf{y}_i^0\}_{i=1}^{N^0}$ are available throughout the entire training. While the images have dimension $\mathbf{x}_i^0 \in \mathbb{R}^{H\times W \times 3}$, with $H$ being the height and $W$ being the width, the segmentation maps are one-hot encoded and follow $\mathbf{y}_i^0 \in \{0, 1\}^{H\times W \times C}$, where $C$ is the number of classes. The target domains $\{\mathcal{D}_1, \mathcal{D}_2, \dots, \mathcal{D}_T\}$ arrive sequentially and without any segmentation maps. Therefore, the dataset of target domain $\mathcal{D}_t$ is given by $X^t = \{\mathbf{x}_j^t\}_{j=1}^{N^t}$, where $N^t$ is again the total number of images. To simplify the notation, we assume that all images of all domains have the same dimensions.   

We divide the style transfer network into an encoder $E$ and a decoder $D$. The output of the encoder is a feature map $E(\mathbf{x}_i) = \mathbf{z}_i \in \mathbb{R}^{H'\times W' \times K'}$, which has $K'$ channels, height $H'$, and width $W'$. The decoder, on the other hand, outputs stylized images $\mathbf{\hat{x}}_i \in \mathbb{R}^{H\times W \times 3}$, that still have the same content as before. Finally, the segmentation model $S$ outputs softmax probability maps $\mathbf{p}_i \in \mathbb{R}^{H\times W \times C}$, which can be converted into one-hot encoded pseudo-labels $\mathbf{\hat{y}}_i \in \{0, 1\}^{H\times W \times C}$. The parameters of the encoder, the decoder, and the segmentation model are denoted by $\boldsymbol{\theta}_{\mathrm{E}}$, $\boldsymbol{\theta}_{\mathrm{D}}$, and $\boldsymbol{\theta}_{\mathrm{S}}$, respectively.  

\subsection{Semantic Segmentation}
Since we have access to the source segmentation maps $\mathbf{y}_i^0$, we can train the segmentation network by computing a pixel-wise cross-entropy (CE) loss of the form 
\begin{align}
\mathcal{L}_{\mathrm{CE}}(\mathbf{x}_i^0, \mathbf{y}_i^0) = \frac{-1}{HWC} \sum_{h=1}^H \sum_{w=1}^W \sum_{c=1}^C \mathbf{y}_{ihwc}^0 \mathrm{log}(\mathbf{p}_{ihwc}^0).
\label{eq:ce_loss}
\end{align}
Although this may lead to a well-performing segmentation model for source data, results typically deteriorate when the model predicts images of an unknown domain. To adapt the segmentation network to a target domain or to counteract forgetting, we transfer images of the source domain $\mathcal{D}_0$ into the style of target domain $\mathcal{D}_t$, without altering the content. To avoid any confusion, we use the notation $\mathbf{\hat{x}}_i^{0\rightarrow t}$ to clearly identify a transferred source image. Given these target-like source images, we can reuse $\mathbf{y}_i^0$ and minimize $\mathcal{L}_{\mathrm{CE}}(\mathbf{\hat{x}}_i^{0\rightarrow t}, \mathbf{y}_i^0)$.  

\subsection{Style Transfer}
Similar to \cite{ACE}, we also build our work on adaptive instance normalization \cite{AdaIN}, summarized below. To transfer an image into an arbitrary style, \cite{AdaIN} first encodes a content image and a style image by taking the respective outputs of layer \textit{relu4} of an ImageNet pre-trained VGG19 \cite{VGG} network. In our case, the content images are from the source domain, while style is drawn from a target domain $t$. The extracted feature maps $\mathbf{z}_i^0$ and $\mathbf{z}_j^t$ of the source and target images, respectively, are then passed to an AdaIN layer. This layer renormalizes the source feature map to have the same mean and standard deviation as the target feature map. Mathematically, this is equivalent to
\begin{equation}
\mathbf{\hat{z}}_i^0 = \mathrm{AdaIN}(\mathbf{z}_i^0, \mathbf{z}_j^t) = \sigma(\mathbf{z}_j^t) \frac{\mathbf{z}_i^0 - \mu(\mathbf{z}_i^0)}{\sigma(\mathbf{z}_i^0)} + \mu(\mathbf{z}_j^t),
\label{eq:plain_adain_layer_op}
\end{equation}
where $\mu(\mathbf{z})$ and $\sigma(\mathbf{z})$ calculate the channel-wise mean and standard deviation.

For the style transfer model to work, the pre-trained encoder is frozen so that only the decoder remains trainable. The loss function minimized by the weights of the decoder $\boldsymbol{\theta}_{\mathrm{D}}$ can be written as 
\begin{align}
\mathcal{L}_{\mathrm{ST}} &= \mathrm{MSE}(E(\mathbf{\hat{x}}_i^{0\rightarrow t}), \mathbf{\hat{z}}_i^0) \nonumber
\\
&+ \lambda \sum_{l=1}^L \bigg[ \mathrm{MSE}\Big( \mu \big (E_l(\mathbf{\hat{x}}_i^{0\rightarrow t}) \big), \mu \big(E_l(\mathbf{x}_j^t) \big) \Big) 
\\ 
&+ \mathrm{MSE} \Big( \sigma \big(E_l(\mathbf{\hat{x}}_i^{0\rightarrow t}) \big ), \sigma \big (E_l(\mathbf{x}_j^t) \big) \Big) \bigg], \nonumber
\label{eq:adain_loss}
\end{align}
where MSE is the mean-squared-error, $E_l()$ represents the output of the $l$-th layer of the encoder, and $\lambda$ is a weighting term \footnote{This notation differs from \cite{AdaIN}. Similar to their publicly available implementation, we use the MSE, while the Euclidean distance is used in \cite{AdaIN}.}. Since \cite{AdaIN} showed that the first two moments capture at least most of the style, a well trained decoder will output an image that depicts the content of the source image, but has the style of the respective target image. 

Although this approach works well for adding a painterly style to an image \cite{AdaIN}, it is not directly suitable for more complex transfers. This is because the AdaIN layer captures the global style of an image. However, for applications like the perception in the real world, different classes in an image have to be treated differently. To overcome this issue, we now condition the computation of the mean and standard deviation on each class $c$ contained in a segmentation mask $\mathbf{y}_i$: 
\begin{align}
\mu_{c}(\mathbf{z}_i, \mathbf{y}_i) &= \frac{1}{\sum_{h,w} \mathbf{y}_{ihwc}} \sum_{h,w} \mathbf{y}_{ihwc} \mathbf{z}_{ihw}
\\
\sigma_{c}(\mathbf{z}_i, \mathbf{y}_i) &= \sqrt{\frac{1}{\sum_{h,w} \mathbf{y}_{ihwc}} \sum_{h,w} \Big (\mathbf{y}_{ihwc} \mathbf{z}_{ihw} - \mu_{c}(\mathbf{z}_i, \mathbf{y}_i) \Big )^2}.
\end{align}
Since we only have access to the segmentation masks of the source domain, we compute the moments of domain $\mathcal{D}_t$ with pseudo-labels $\mathbf{\hat{y}}_j^t$. Please note that we circumvent a possible mismatch in resolution between the segmentation mask and the feature map by simply resizing the height and width of mask $\mathbf{y}_i^0$ or $\mathbf{\hat{y}}_j^t$ to the correct size. 
Then, for each class $c$ contained in the source segmentation mask $\mathbf{y}_i^0$, the class-conditional AdaIN layer renormalizes every region of $\mathbf{z}_i^0$ belonging to $c$ by using class-specific moments. This results in
\begin{equation}
\mathbf{\hat{z}}_i^0 = \sum_c \mathbf{y}_{ic}^0 \odot \bigg( \sigma_c(\mathbf{z}_j^t, \mathbf{\hat{y}}_j^t) \frac{\mathbf{z}_i^0 - \mu_c(\mathbf{z}_i^0, \mathbf{y}_i^0)}{\sigma_c(\mathbf{z}_i^0, \mathbf{y}_i^0)} + \mu_c(\mathbf{z}_j^t, \mathbf{\hat{y}}_j^t)\bigg ),
\label{eq:spatial_adain_layer_op}
\end{equation}
with $\odot$ being an element-wise multiplication. Since we use a class-conditional AdaIN layer, we slightly adapt the training loss of the decoder to: 
\begin{align}
& \mathcal{L}_{\mathrm{ST}} = \mathrm{MSE}(E(\mathbf{\hat{x}}_i^{0\rightarrow t}), \, \mathbf{\hat{z}}_i^0) \nonumber
\\ 
+ &\frac{\lambda}{C} \sum_{l=1}^L \sum_c \bigg [  \mathrm{MSE} \Big( \mu_c \big( E_l(\mathbf{\hat{x}}_i^{0 \rightarrow t}), \mathbf{y}_i^0 \big), \, \mu_c \big( E_l(\mathbf{x}_j^t), \mathbf{\hat{y}}_j^t \big) \Big) \nonumber
\\
+ &\mathrm{MSE} \Big( \sigma_c \big( E_l(\mathbf{\hat{x}}_i^{0 \rightarrow t}), \mathbf{y}_i^0 \big), \, \sigma_c(E_l(\mathbf{x}_j^t), \mathbf{\hat{y}}_j^t) \Big) \bigg ]. 
\label{eq:spatial_adain_loss}
\end{align}
Clearly, there is the possibility that the source and the target images do not contain the exact same classes. However, to compute \eqref{eq:spatial_adain_layer_op} and \eqref{eq:spatial_adain_loss}, one can also draw the required moments from different images, which we do during training. This also effectively enlarges the variety of styles.

\begin{figure}
\def\svgwidth{250pt}
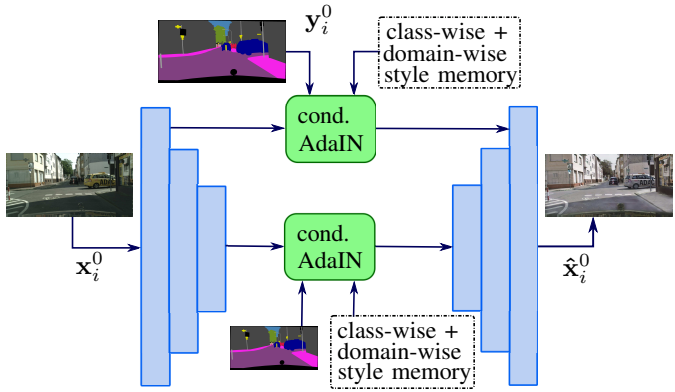
\caption{Overview of our style transfer network. It uses two class-conditional AdaIN layers that query the required target moments from a style memory.}
\label{fig:adain_model}
\end{figure}

Although a class-conditional AdaIN layer helps to capture the modes in a target image much better, the generated outputs still suffer from the same artifacts as before. To reduce the artifacts, we further add a skip connection between the encoder and decoder and equip it with another conditional AdaIN layer. Since the feature map used for the skip connection still has the same resolution as the input image, the newly inserted layer helps to transfer the style of small objects and between class boundaries more accurately. An overview of our style transfer network is illustrated in Fig. \ref{fig:adain_model}.

\subsection{Style Memory} 
One problem that usually occurs when a model is trained sequentially is catastrophic forgetting. This causes the performance of the model for domain $\mathcal{D}_{t-1}$ to decrease again when it is trained on the next target domain $\mathcal{D}_{t}$. A very effective technique to counteract forgetting is rehearsal \cite{Remind, GenReplay}. Since we do not consider the supervised case, we follow the approach of \cite{ACE} and only save encountered styles, i.e., we save the class-wise means and the standard deviations. Since the input images for urban scene segmentation are usually of high resolution, this drastically reduces the required memory. Nevertheless, if the available memory is even more limited, it is also possible to store only a certain percentage of moments. Alternatively, one can also estimate a Gaussian distribution from the extracted moments of each class $c$ of domain $\mathcal{D}_t$. Subsequently, the required moments can be sampled from the estimated distribution. Depending on the size of the dataset, this may again heavily decrease the required storage size. 

An overview of the training procedure of our complete framework is described in Algorithm \ref{alg:training}. 
\begin{algorithm}
	\caption{CACE Training Procedure}
	\begin{algorithmic}[1]
	\Require{Source pre-trained $\boldsymbol{\theta}_{S}$, ImageNet pre-trained $\boldsymbol{\theta}_{E}$}
	\State Initialize empty moment dictionary $M$
		\For {$t = 1, 2, \dots, T$}
		\State Create pseudo-labels $\mathbf{\hat{y}}^t$ for target domain $\mathcal{D}_t$
		\State Extract class-wise moments for domain $\mathcal{D}_t$ using $\mathbf{\hat{y}}^t$
		\State Save class-wise moments in $M$
		\For {$i = 1, 2, \dots, I$} 		\Comment{Train style transfer model}
			\State Sample mini-batch from $X^0$
			\State Draw moments from $M$ according to Sec. \ref{Sec_implementation}
			\State Calculate $\mathcal{L}_{\mathrm{ST}}$ \eqref{eq:spatial_adain_loss} and update $\boldsymbol{\theta}_{D}$
		\EndFor
		
		\For {$j = 1, 2, \dots, J$} \Comment{Train segmentation model}
			\State Sample mini-batch from $X^0$
			\State Transfer samples according to Sec. \ref{Sec_implementation}
			\State Calculate $\mathcal{L}_{\mathrm{CE}}$   and update $\boldsymbol{\theta}_{S}$
		\EndFor
		
		\EndFor
	\end{algorithmic} 
	\label{alg:training}
\end{algorithm}


\section{Experiments}
\subsection{Experimental Settings}
\subsubsection{Datasets and Metrics} 
We evaluate our approach using two different domain sequences that roughly capture varying illuminations as well as adverse weather conditions. While the first sequence consists of 5 real-world domains, the second one contains 11 synthetic domains. The real-world sequence uses Cityscapes (CS) \cite{Cityscapes} as labeled source domain. Cityscapes contains 2975 training and 500 validation images at a resolution of $1024 \times 2048$. The real-world target domains originate from the ACDC \cite{ACDC} dataset and arrive in the following order: \textit{Fog}, \textit{Night}, \textit{Rain}, and finally \textit{Snow}. Every domain in the ACDC dataset consists of approximately 400 training and 100 validation images with a resolution of $1080 \times 1920$. The synthetic sequence is derived from SYNTHIA-SEQ, a subset of the Synthia dataset \cite{SYNTHIA}. In this case, we choose the \textit{Old European Town} subset that contains 11 domains with roughly 1000 images of resolution $760 \times 1280$ per domain. We split each domain equally in one training set and one validation set, and use the domain \textit{Dawn} as labeled source domain. The target domains, on the other hand, can be seen in Table \ref{tab:main_synthia}, arriving in order from left to right. While we use the same 19 classes as in \cite{ACDC} for the real-world sequence, we use 13 classes for the synthetic one. As a performance metric, we use the mean intersection-over-union (mIoU), where we show the results for each individual domain after completing the training on the last domain. In addition, we also compute the mean of all domain-wise mIoUs (mean mIoU).

\subsubsection{Network architecture} 
We use the standard framework for UDA in semantic segmentation \cite{AdaptSegNet}, which deploys the DeepLab-V2 \cite{Deeplabv2} framework with a ResNet-101 as the feature extractor. The weights of the ResNet-101 are pre-trained on ImageNet and all batch normalization layers are frozen during training. For the style transfer network, we follow \cite{AdaIN} and use a VGG19-based encoder-decoder architecture, where the encoder is initialized with weights pre-trained on ImageNet as well. Note that the parameters of the encoder are frozen during the entire training.

\subsubsection{Implementation Details} \label{Sec_implementation}
The network is trained using SGD with momentum of $0.9$, a constant learning rate set to $2.5\times10^{-4}$ and weight decay of $5\times10^{-4}$. During training, we use random horizontal flipping. Furthermore, we rescale the shorter size of an image to size $640$ for the real sequence, and size $760$ for the synthetic one,  before randomly cropping the image to size $512\times1024$. We train our model using batch size 2, where one sample is transferred into the style of the current target domain $\mathcal{D}_t$ while the other is randomly chosen to be either from the original source domain $\mathcal{D}_0$ or transferred into a style of a previous target domain {$\mathcal{D}_1, \dots, \mathcal{D}_{t-1}$}. While we pre-train the segmentation model for 100k iterations on the real source domain, it is only pre-trained for 50k iterations on the much simpler synthetic source domain. In both cases, we continue training the segmentation network for 10k iterations ($J$) every time a new target domain arrives. The style transfer network is initially trained for 30k iterations using Adam optimizer with learning rate set to $1\times10^{-4}$. The data during this period is sampled from $\mathcal{D}_0$ and $\mathcal{D}_1$. Further, we fine tune the style transfer model for another 5k iterations every time the target domain changes. To prevent it from forgetting, half of the batch uses moments from the current target domain $t$, while the other half randomly draws its moment from the previous domains $[1, \dots, t-1]$.

\subsubsection{Baselines}
Since it is reported in \cite{ACDC} that for Cityscapes $\rightarrow$ ACDC, most of the current adversarial learning or self-training based approaches proposed for unsupervised domain adaptation could not improve upon the source only baseline, we do not consider such methods in our even more challenging setting. However, \cite{ACDC} also found that Fourier domain adaptation (FDA) \cite{FDA} could effectively increase the segmentation results. Therefore, we use their light-weight style transfer method as a baseline to evaluate our approach. Note that \cite{ACDC} also evaluated BDL \cite{Bidirectional}, which performed significantly worse than FDA despite using the CycleGAN framework for style transfer. We also reimplement the ACE framework with the minor difference that the style transfer network is not jointly trained with the segmentation model to save some memory. Further, we also compare to only using their memory efficient implementation of the AdaIN model in our framework. Finally, we carefully investigate the impact of color jittering.  

\subsection{Results and Discussion}
We start our investigations with the challenging real-world sequence. To illustrate the effects of the augmentation color jittering, we use two different settings, which differ in whether or not color jittering was used for pre-training the source model. The results without color jittering during pre-training are depicted in Table \ref{tab:main_real_no_col}. Clearly, all methods improve upon the source only baseline by at least $3\%$ in terms of mean mIoU. However, if the source model is further trained with color jittering for the same number of update steps, the improvement decreases. When using the Fourier based style transfer, the results are now even slightly worse. Our approach, on the other hand, does even improve upon color jittering, when we sample the required moments from an estimated Gaussian distribution (Ours, d). When we use moments extracted from target samples (Ours, s), the performance increases further by $1.5\%$ in terms of mean mIoU, with the highest increase observed for the \textit{Night} domain. Since this domain contains yellowish as well as whitish images due to varying street lighting, the estimate of a unimodal Gaussian distribution is too simple.
\begin{table}[h]
\caption{mIoU after adapting the model to the last target domain. The initial model was pre-trained on Cityscapes (CS) without color jittering.}
\begin{center}
\begin{tabular}{ l|ccccc|c }
  \hline
  & CS & Fog & Night & Rain & Snow & mean mIoU \\
  \hline
 Source & 71.5 & 52.5 &  19.0 & 37.4 & 41.8 & 44.4 \\
 Jitter & $\mathbf{72.6}$ & 60.1 & 22.1 & 42.0 & 44.3 & 48.2 \\
 AdaIN & 69.0 & 59.1 & 23.9 & 45.4 & 49.1 & 49.3 \\
 ACE & 70.3 & 57.2 & 23.8 &  46.4 & 47.3 & 49.0 \\
 FDA & 71.5 & 55.6 & 22.3 & 43.4 & 44.0 & 47.4\\
  \hline
 Ours, d & 68. & $\mathbf{62.3}$ & 21.8 & 46.3 & 48.9 & 49.5 \\
 Ours, s & 69.1 & 62.1 & $\mathbf{25.3}$ & $\mathbf{47.6}$ & $\mathbf{51.1}$ & $\mathbf{51.0}$ \\
  \hline
\end{tabular}
\end{center}
\label{tab:main_real_no_col}
\end{table}
\begin{table}[h]
\caption{mIoU after adapting the model to the last target domain. The initial model was pre-trained on Cityscapes (CS) with color jittering.}
\begin{center}
\begin{tabular}{ l|ccccc|c }
  \hline
  & CS & Fog & Night & Rain & Snow & mean mIoU \\
  \hline
 Source & 70.7 & 62.9 & 22.6 & 45.0 & 46.6 & 49.6 \\
 Jitter & $\mathbf{72.0}$ & 61.7 & 22.8 & 44.4 & 44.4 & 49.1 \\
 AdaIN & 69.9 & 59.0 & 24.1 & 45.9 & 50.5 & 49.9 \\
 ACE & 70.4 & 58.5 & 26.3 & 47.6 & 50.4 & 50.6\\
 FDA & 71.7 & 59.7 & 24.7 & 45.2 & 42.5 & 48.8 \\
  \hline
 Ours, d & 67.7 & 65.1 & 25.9 & 47.5 & 49.6 & 51.2 \\
 Ours, s & 69.1 & $\mathbf{65.7}$ & $\mathbf{29.7}$ & $\mathbf{47.7}$ & $\mathbf{50.6}$ & $\mathbf{52.6}$ \\
  \hline
\end{tabular}
\end{center}
\label{tab:main_real_from_col}
\end{table}
Since the previously mentioned table indicates that color jittering is already a good baseline, we now examine the effects when the model is directly pre-trained with color jittering. The results for this scenario are shown in Table \ref{tab:main_real_from_col}. In this case, the source only baseline improves dramatically and is now comparable or only slightly worse to the other non-class-specific transfer methods. In contrast, our approach even profits from the stronger baseline model, improving the results by about $1.6\%$ points in terms of mean mIoU in both cases. This is due to the better approximation of the class-wise target moments, which are now computed from more accurate pseudo-labels.

A much larger performance increase can be seen for the sequence consisting of synthetic domains only. Following the same evaluation strategy as before, one can see in Table \ref{tab:main_synthia} that now all methods are clearly superior to color jittering, even when it is directly used during pre-training. Nonetheless, our class-conditioned style transfer again outperforms all other methods by a large margin. This applies not only when the moments are drawn from samples (Ours, s), but also when we again estimate a Gaussian distribution (Ours, d) to increase the memory efficiency.
\begin{table*}[h]
\caption{mIoU of the final model for each synthetic domain. Left: no color jittering during the source pre-training. Right: using color jittering during pre-training.}
\footnotesize
\setlength\tabcolsep{3.0pt}
\begin{center}
\begin{tabular}{ l|ccccccccccc|>{\columncolor[gray]{0.95}}c|ccccccccccc|>{\columncolor[gray]{0.95}}c }
 & \multicolumn{12}{c}{\small{No color jitter during pre-training}} & \multicolumn{12}{c}{\small{Using color jitter during pre-training}} \\
  \hline
  & \rotatebox[origin=l]{90}{Dawn } & \rotatebox[origin=l]{90}{Fall } & \rotatebox[origin=l]{90}{Fog } & \rotatebox[origin=l]{90}{Night } & \rotatebox[origin=l]{90}{Rainnight } & \rotatebox[origin=l]{90}{Softrain } & \rotatebox[origin=l]{90}{Spring } & \rotatebox[origin=l]{90}{Summer } & \rotatebox[origin=l]{90}{Sunset } & \rotatebox[origin=l]{90}{Winter } & \rotatebox[origin=l]{90}{Winternight } & \rotatebox[origin=l]{90}{mean mIoU } & \rotatebox[origin=l]{90}{Dawn } & \rotatebox[origin=l]{90}{Fall } & \rotatebox[origin=l]{90}{Fog } & \rotatebox[origin=l]{90}{Night } & \rotatebox[origin=l]{90}{Rainnight } & \rotatebox[origin=l]{90}{Softrain } & \rotatebox[origin=l]{90}{Spring } & \rotatebox[origin=l]{90}{Summer } & \rotatebox[origin=l]{90}{Sunset } & \rotatebox[origin=l]{90}{Winter } & \rotatebox[origin=l]{90}{Winternight } & \rotatebox[origin=l]{90}{mean mIoU } \\
  \hline
 Source & 70.5 & 56.5 & 58.6 & 53.7 & 26.2 & 37.1 & 56.6 & 47.6 & 66.0 & 40.0 & 40.7 & 50.3 & 70.3 & 62.4 & 64.0 & 62.3 & 29.2 & 38.7 & 57.5 & 53.8 & 70.6 & 40.4 & 49.7 & 54.5 \\
 Jitter & 71.6 & 62.0 & 64.0 & 59.8 & 29.0 & 41.1 & 57.9 & 55.4 & 71.7 & 44.5 & 50.5 & 55.2 & 71.6 & 63.7 & 65.7 & 61.9 & 29.4 & 39.5 & 59.2 & 56.0 & 71.5 & 42.8 & 51.7 & 55.7 \\
 AdaIN & 71.3 & 64.9 & 67.1 & 58.8 & 34.1 & 46.9 & 61.6 & 61.1 & 70.8 & 47.3 & 50.5 & 57.7 & 71.0 & 65.5 & 67.3 & 63.7 & 35.3 & 46.7 & 62.4 & 59.9 & 71.1 & 45.7 & 54.4 & 58.5 \\
 ACE & 71.6 & 65.4 & 67.8 & 61.4 & 37.1 & 48.7 & 63.5 & 61.3 & 71.2 & 46.5 & 51.5 & 58.7 & 71.2 & 66.1 & 67.4 & 61.5 & 38.1 & 49.5 & 63.0 & 62.0 & 71.4 & 44.4 & 52.2 & 58.8  \\  
 FDA & 71.8 & 65.1 & 66.3 & 61.4 & 33.2 & 39.6 & 63.8 & 66.5 & 70.8 & 47.7 & 51.3 & 58.0 & 71.2 & 64.4 & 65.5 & 61.3 & 36.1 & 41.5 & 63.2 & 65.5 & 71.0 & 46.5 & 52.6 & 58.1 \\
 \hline
  Ours, d & 71.4 & 67.7 & 67.2 & 64.1 & 42.6 & 49.5 & 65.5 & 67.5 & 71.3 & 51.2 & 57.7 & 61.4 & 71.1 & 68.3 & 67.8 & 66.2 & 43.9 & 51.2 & 65.9 & 67.5 & 71.5 & 52.2 & 58.5 & 62.2 \\
  Ours, s & 71.4 & 69.0 & 68.5 & 66.5 & 49.1 & 57.9 & 67.3 & 69.7 & 71.3 & 53.7 & 59.1 & 63.9 & 71.2 & 69.3 & 68.5 & 67.9 & 49.1 & 58.1 & 67.1 & 69.7 & 71.7 & 54.3 & 60.0 & 64.3 \\
  \hline
\end{tabular}
\end{center}
\label{tab:main_synthia}
\end{table*}


\subsection{Ablation Studies}
Unless otherwise stated, we use the real-world sequence without color jittering (CJ) during pre-training to carefully examine our approach.
\begin{table}[h]
\caption{Results when class-wise target moments are extracted using pseudo-labels (PL) and real target labels (RL) for: pre-training the model with and without color jittering (CJ).}
\setlength\tabcolsep{4.5pt}
\begin{center}
\begin{tabular}{ l|ccccc|c }
  \hline
  & CS & Fog & Night & Rain & Snow & mean mIoU \\
  \hline
 No CJ, from PL & 69.1 & 62.1 & 25.3 & 47.6 & 51.1 & 51.0 \\
 No CJ, from RL & $\mathbf{70.2}$ & $\mathbf{62.4}$ & $\mathbf{28.3}$ & $\mathbf{49.2}$ & $\mathbf{54.1}$ & $\mathbf{52.8}$ \\
  \hline
 CJ, from PL & 69.1 & $\mathbf{65.7}$ & $\mathbf{29.7}$ & 47.7 & 50.6 & 52.6 \\
 CJ, from RL & $\mathbf{69.3}$ & 64.6 & 29.1 & $\mathbf{49.3}$ & $\mathbf{53.6}$ & $\mathbf{53.2}$ \\
  \hline
\end{tabular}
\end{center}
\label{tab:ablation_pl}
\end{table}
Since we use pseudo-labels (PL) to estimate the corresponding class-wise target moments, we now study the effect, when one could use real target labels (RL) to extract the moments. Again, we do the evaluation for a model that was pre-trained with and without color jittering (CJ) on the source domain. As illustrated by Table \ref{tab:ablation_pl}, the results increase with the quality of the target moments. However, it also shows that even using unfiltered pseudo-labels as an approximation does not drastically degrade the results.  

\begin{figure}[b]
\def\svgwidth{250pt}
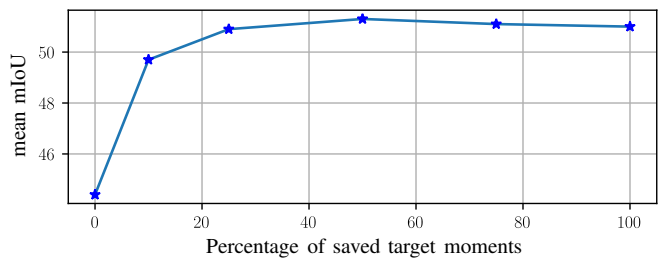
\caption{Performance compared to the percentage of saved moments.}
\label{fig:ablation_memory}
\end{figure}

Next, we examine the impact of reducing the number of stored moments to save some memory. As Fig. \ref{fig:ablation_memory} shows, the results do not deteriorate significantly if we only save $25\%$ of all target moments. If memory is a real bottleneck and one can only store $10\%$, one may consider estimating a Gaussian distribution, which yields comparable results here while being more memory efficient.

To investigate whether catastrophic forgetting is a real problem, we conducted an experiment in which we did not replay source images in the style of previously seen domain. In this case, the results decreased by $2.2\%$ in terms of mean mIoU. When we did not even use original source samples, but transferred both images of our batch to the style of the current target domain, the mean mIoU decreased from $51.$ to $45.4$.


\section{Conclusion}
In this work, we propose a simple framework that uses light-weight style transfer to adapt a pre-trained source model to a sequence of unlabeled target domains while preventing forgetting. Unlike previous work, we condition the style transfer network, which is based on adaptive instance normalization, on each class. Our approach outperforms comparable methods on a challenging real-world sequence as well as a synthetic sequence.


\bibliographystyle{IEEEtran}
\bibliography{egbib}

\begin{thebibliography}{10}
\providecommand{\url}[1]{#1}
\csname url@samestyle\endcsname
\providecommand{\newblock}{\relax}
\providecommand{\bibinfo}[2]{#2}
\providecommand{\BIBentrySTDinterwordspacing}{\spaceskip=0pt\relax}
\providecommand{\BIBentryALTinterwordstretchfactor}{4}
\providecommand{\BIBentryALTinterwordspacing}{\spaceskip=\fontdimen2\font plus
\BIBentryALTinterwordstretchfactor\fontdimen3\font minus
  \fontdimen4\font\relax}
\providecommand{\BIBforeignlanguage}[2]{{%
\expandafter\ifx\csname l@#1\endcsname\relax
\typeout{** WARNING: IEEEtran.bst: No hyphenation pattern has been}%
\typeout{** loaded for the language `#1'. Using the pattern for}%
\typeout{** the default language instead.}%
\else
\language=\csname l@#1\endcsname
\fi
#2}}
\providecommand{\BIBdecl}{\relax}
\BIBdecl

\bibitem{Deeplabv2}
L.-C. Chen, G.~Papandreou, I.~Kokkinos, K.~Murphy, and A.~L. Yuille, ``Deeplab:
  Semantic image segmentation with deep convolutional nets, atrous convolution,
  and fully connected crfs,'' \emph{IEEE transactions on pattern analysis and
  machine intelligence}, vol.~40, no.~4, pp. 834--848, 2017.

\bibitem{DeepLabv3+}
L.-C. Chen, Y.~Zhu, G.~Papandreou, F.~Schroff, and H.~Adam, ``Encoder-decoder
  with atrous separable convolution for semantic image segmentation,'' in
  \emph{Proceedings of the European conference on computer vision (ECCV)},
  2018, pp. 801--818.

\bibitem{HRNet}
K.~Sun, Y.~Zhao, B.~Jiang, T.~Cheng, B.~Xiao, D.~Liu, Y.~Mu, X.~Wang, W.~Liu,
  and J.~Wang, ``High-resolution representations for labeling pixels and
  regions,'' \emph{arXiv preprint arXiv:1904.04514}, 2019.

\bibitem{Cityscapes}
M.~Cordts, M.~Omran, S.~Ramos, T.~Rehfeld, M.~Enzweiler, R.~Benenson,
  U.~Franke, S.~Roth, and B.~Schiele, ``The cityscapes dataset for semantic
  urban scene understanding,'' in \emph{Proceedings of the IEEE conference on
  computer vision and pattern recognition}, 2016, pp. 3213--3223.

\bibitem{CYCADA}
J.~Hoffman, E.~Tzeng, T.~Park, J.-Y. Zhu, P.~Isola, K.~Saenko, A.~Efros, and
  T.~Darrell, ``Cycada: Cycle-consistent adversarial domain adaptation,'' in
  \emph{International conference on machine learning}.\hskip 1em plus 0.5em
  minus 0.4em\relax PMLR, 2018, pp. 1989--1998.

\bibitem{Bidirectional}
Y.~Li, L.~Yuan, and N.~Vasconcelos, ``Bidirectional learning for domain
  adaptation of semantic segmentation,'' in \emph{Proceedings of the IEEE
  Conference on Computer Vision and Pattern Recognition}, 2019, pp. 6936--6945.

\bibitem{FDA}
Y.~Yang and S.~Soatto, ``Fda: Fourier domain adaptation for semantic
  segmentation,'' in \emph{Proceedings of the IEEE/CVF Conference on Computer
  Vision and Pattern Recognition}, 2020, pp. 4085--4095.

\bibitem{ACE}
Z.~Wu, X.~Wang, J.~E. Gonzalez, T.~Goldstein, and L.~S. Davis, ``Ace: adapting
  to changing environments for semantic segmentation,'' in \emph{Proceedings of
  the IEEE International Conference on Computer Vision}, 2019, pp. 2121--2130.

\bibitem{FCNs}
J.~Hoffman, D.~Wang, F.~Yu, and T.~Darrell, ``Fcns in the wild: Pixel-level
  adversarial and constraint-based adaptation,'' \emph{arXiv preprint
  arXiv:1612.02649}, 2016.

\bibitem{ActivationMatching}
H.~Huang, Q.~Huang, and P.~Krahenbuhl, ``Domain transfer through deep
  activation matching,'' in \emph{Proceedings of the European Conference on
  Computer Vision (ECCV)}, 2018, pp. 590--605.

\bibitem{CAG}
Q.~Zhang, J.~Zhang, W.~Liu, and D.~Tao, ``Category anchor-guided unsupervised
  domain adaptation for semantic segmentation,'' in \emph{Advances in Neural
  Information Processing Systems}, 2019, pp. 435--445.

\bibitem{CLST}
R.~A. Marsden, A.~Bartler, M.~D{\"o}bler, and B.~Yang, ``Contrastive learning
  and self-training for unsupervised domain adaptation in semantic
  segmentation,'' \emph{arXiv preprint arXiv:2105.02001}, 2021.

\bibitem{AdaptSegNet}
Y.-H. Tsai, W.-C. Hung, S.~Schulter, K.~Sohn, M.-H. Yang, and M.~Chandraker,
  ``Learning to adapt structured output space for semantic segmentation,'' in
  \emph{Proceedings of the IEEE Conference on Computer Vision and Pattern
  Recognition}, 2018, pp. 7472--7481.

\bibitem{ADVENT}
T.-H. Vu, H.~Jain, M.~Bucher, M.~Cord, and P.~P{\'e}rez, ``Advent: Adversarial
  entropy minimization for domain adaptation in semantic segmentation,'' in
  \emph{Proceedings of the IEEE conference on computer vision and pattern
  recognition}, 2019, pp. 2517--2526.

\bibitem{PatchAlign}
Y.-H. Tsai, K.~Sohn, S.~Schulter, and M.~Chandraker, ``Domain adaptation for
  structured output via discriminative patch representations,'' in
  \emph{Proceedings of the IEEE International Conference on Computer Vision},
  2019, pp. 1456--1465.

\bibitem{CLAN}
Y.~Luo, L.~Zheng, T.~Guan, J.~Yu, and Y.~Yang, ``Taking a closer look at domain
  shift: Category-level adversaries for semantics consistent domain
  adaptation,'' in \emph{Proceedings of the IEEE Conference on Computer Vision
  and Pattern Recognition}, 2019, pp. 2507--2516.

\bibitem{DACS}
W.~Tranheden, V.~Olsson, J.~Pinto, and L.~Svensson, ``Dacs: Domain adaptation
  via cross-domain mixed sampling,'' in \emph{Proceedings of the IEEE/CVF
  Winter Conference on Applications of Computer Vision}, 2021, pp. 1379--1389.

\bibitem{CF}
M.~McCloskey and N.~J. Cohen, ``Catastrophic interference in connectionist
  networks: The sequential learning problem,'' in \emph{Psychology of learning
  and motivation}.\hskip 1em plus 0.5em minus 0.4em\relax Elsevier, 1989,
  vol.~24, pp. 109--165.

\bibitem{AdaIN}
X.~Huang and S.~Belongie, ``Arbitrary style transfer in real-time with adaptive
  instance normalization,'' in \emph{Proceedings of the IEEE International
  Conference on Computer Vision}, 2017, pp. 1501--1510.

\bibitem{GAN}
I.~Goodfellow, J.~Pouget-Abadie, M.~Mirza, B.~Xu, D.~Warde-Farley, S.~Ozair,
  A.~Courville, and Y.~Bengio, ``Generative adversarial nets,'' in
  \emph{Advances in Neural Information Processing Systems}, Z.~Ghahramani,
  M.~Welling, C.~Cortes, N.~Lawrence, and K.~Q. Weinberger, Eds.,
  vol.~27.\hskip 1em plus 0.5em minus 0.4em\relax Curran Associates, Inc.,
  2014, pp. 2672--2680.

\bibitem{CycleGAN}
J.-Y. Zhu, T.~Park, P.~Isola, and A.~A. Efros, ``Unpaired image-to-image
  translation using cycle-consistent adversarial networks,'' in
  \emph{Proceedings of the IEEE international conference on computer vision},
  2017, pp. 2223--2232.

\bibitem{UNIT}
M.-Y. Liu, T.~Breuel, and J.~Kautz, ``Unsupervised image-to-image translation
  networks,'' \emph{Advances in neural information processing systems},
  vol.~30, 2017.

\bibitem{MUNIT}
X.~Huang, M.-Y. Liu, S.~Belongie, and J.~Kautz, ``Multimodal unsupervised
  image-to-image translation,'' in \emph{Proceedings of the European conference
  on computer vision (ECCV)}, 2018, pp. 172--189.

\bibitem{CUT}
T.~Park, A.~A. Efros, R.~Zhang, and J.-Y. Zhu, ``Contrastive learning for
  unpaired image-to-image translation,'' in \emph{European Conference on
  Computer Vision}.\hskip 1em plus 0.5em minus 0.4em\relax Springer, 2020, pp.
  319--345.

\bibitem{Gram}
L.~A. Gatys, A.~S. Ecker, and M.~Bethge, ``Image style transfer using
  convolutional neural networks,'' in \emph{Proceedings of the IEEE conference
  on computer vision and pattern recognition}, 2016, pp. 2414--2423.

\bibitem{PerceptualLoss}
J.~Johnson, A.~Alahi, and L.~Fei-Fei, ``Perceptual losses for real-time style
  transfer and super-resolution,'' in \emph{European conference on computer
  vision}.\hskip 1em plus 0.5em minus 0.4em\relax Springer, 2016, pp. 694--711.

\bibitem{ST_MMD}
Y.~Li, N.~Wang, J.~Liu, and X.~Hou, ``Demystifying neural style transfer,''
  \emph{arXiv preprint arXiv:1701.01036}, 2017.

\bibitem{ST_CMD}
N.~Kalischek, J.~D. Wegner, and K.~Schindler, ``In the light of feature
  distributions: moment matching for neural style transfer,'' in
  \emph{Proceedings of the IEEE/CVF Conference on Computer Vision and Pattern
  Recognition}, 2021, pp. 9382--9391.

\bibitem{MMD}
D.~Sejdinovic, B.~Sriperumbudur, A.~Gretton, and K.~Fukumizu, ``Equivalence of
  distance-based and rkhs-based statistics in hypothesis testing,'' \emph{The
  Annals of Statistics}, pp. 2263--2291, 2013.

\bibitem{CMD}
W.~Zellinger, B.~A. Moser, T.~Grubinger, E.~Lughofer, T.~Natschl{\"a}ger, and
  S.~Saminger-Platz, ``Robust unsupervised domain adaptation for neural
  networks via moment alignment,'' \emph{Information Sciences}, vol. 483, pp.
  174--191, 2019.

\bibitem{Proxy}
T.~Shen, D.~Gong, W.~Zhang, C.~Shen, and T.~Mei, ``Regularizing proxies with
  multi-adversarial training for unsupervised domain-adaptive semantic
  segmentation,'' \emph{arXiv preprint arXiv:1907.12282}, 2019.

\bibitem{NoMoreDiscrimination}
Y.-H. Chen, W.-Y. Chen, Y.-T. Chen, B.-C. Tsai, Y.-C. Frank~Wang, and M.~Sun,
  ``No more discrimination: Cross city adaptation of road scene segmenters,''
  in \emph{Proceedings of the IEEE International Conference on Computer
  Vision}, 2017, pp. 1992--2001.

\bibitem{SIM}
Z.~Wang, M.~Yu, Y.~Wei, R.~Feris, J.~Xiong, W.-m. Hwu, T.~S. Huang, and H.~Shi,
  ``Differential treatment for stuff and things: A simple unsupervised domain
  adaptation method for semantic segmentation,'' in \emph{Proceedings of the
  IEEE/CVF Conference on Computer Vision and Pattern Recognition}, 2020, pp.
  12\,635--12\,644.

\bibitem{MRNet}
Z.~Zheng and Y.~Yang, ``Unsupervised scene adaptation with memory
  regularization in vivo,'' \emph{arXiv preprint arXiv:1912.11164}, 2019.

\bibitem{IAST}
K.~Mei, C.~Zhu, J.~Zou, and S.~Zhang, ``Instance adaptive self-training for
  unsupervised domain adaptation,'' \emph{arXiv preprint arXiv:2008.12197},
  2020.

\bibitem{CBST}
Y.~Zou, Z.~Yu, B.~Vijaya~Kumar, and J.~Wang, ``Unsupervised domain adaptation
  for semantic segmentation via class-balanced self-training,'' in
  \emph{Proceedings of the European conference on computer vision (ECCV)},
  2018, pp. 289--305.

\bibitem{CCM}
G.~Li, G.~Kang, W.~Liu, Y.~Wei, and Y.~Yang, ``Content-consistent matching for
  domain adaptive semantic segmentation,'' in \emph{European Conference on
  Computer Vision}.\hskip 1em plus 0.5em minus 0.4em\relax Springer, 2020, pp.
  440--456.

\bibitem{MaxSquares}
M.~Chen, H.~Xue, and D.~Cai, ``Domain adaptation for semantic segmentation with
  maximum squares loss,'' in \emph{Proceedings of the IEEE International
  Conference on Computer Vision}, 2019, pp. 2090--2099.

\bibitem{RectifyingPL}
Z.~Zheng and Y.~Yang, ``Rectifying pseudo label learning via uncertainty
  estimation for domain adaptive semantic segmentation,'' \emph{International
  Journal of Computer Vision}, pp. 1--15, 2021.

\bibitem{EWC}
J.~Kirkpatrick, R.~Pascanu, N.~Rabinowitz, J.~Veness, G.~Desjardins, A.~A.
  Rusu, K.~Milan, J.~Quan, T.~Ramalho, A.~Grabska-Barwinska \emph{et~al.},
  ``Overcoming catastrophic forgetting in neural networks,'' \emph{Proceedings
  of the national academy of sciences}, vol. 114, no.~13, pp. 3521--3526, 2017.

\bibitem{SI}
F.~Zenke, B.~Poole, and S.~Ganguli, ``Continual learning through synaptic
  intelligence,'' in \emph{International Conference on Machine Learning}.\hskip
  1em plus 0.5em minus 0.4em\relax PMLR, 2017, pp. 3987--3995.

\bibitem{RI}
A.~Chaudhry, P.~K. Dokania, T.~Ajanthan, and P.~H.~S. Torr, ``Riemannian walk
  for incremental learning: Understanding forgetting and intransigence,'' in
  \emph{Proceedings of the European Conference on Computer Vision (ECCV)},
  September 2018.

\bibitem{GenReplay}
H.~Shin, J.~K. Lee, J.~Kim, and J.~Kim, ``Continual learning with deep
  generative replay,'' \emph{Advances in neural information processing
  systems}, vol.~30, 2017.

\bibitem{Remind}
T.~L. Hayes, K.~Kafle, R.~Shrestha, M.~Acharya, and C.~Kanan, ``Remind your
  neural network to prevent catastrophic forgetting,'' in \emph{European
  Conference on Computer Vision}.\hskip 1em plus 0.5em minus 0.4em\relax
  Springer, 2020, pp. 466--483.

\bibitem{PNN}
A.~A. Rusu, N.~C. Rabinowitz, G.~Desjardins, H.~Soyer, J.~Kirkpatrick,
  K.~Kavukcuoglu, R.~Pascanu, and R.~Hadsell, ``Progressive neural networks,''
  \emph{arXiv preprint arXiv:1606.04671}, 2016.

\bibitem{DEN}
J.~Yoon, E.~Yang, J.~Lee, and S.~J. Hwang, ``Lifelong learning with dynamically
  expandable networks,'' \emph{arXiv preprint arXiv:1708.01547}, 2017.

\bibitem{ETM}
J.~Kim, S.-M. Yoo, G.-M. Park, and J.-H. Kim, ``Continual unsupervised domain
  adaptation for semantic segmentation,'' \emph{arXiv preprint
  arXiv:2010.09236}, 2020.

\bibitem{VGG}
K.~Simonyan and A.~Zisserman, ``Very deep convolutional networks for
  large-scale image recognition,'' \emph{arXiv preprint arXiv:1409.1556}, 2014.

\bibitem{ACDC}
C.~Sakaridis, D.~Dai, and L.~Van~Gool, ``Acdc: The adverse conditions dataset
  with correspondences for semantic driving scene understanding,'' in
  \emph{Proceedings of the IEEE/CVF International Conference on Computer
  Vision}, 2021, pp. 10\,765--10\,775.

\bibitem{SYNTHIA}
G.~Ros, L.~Sellart, J.~Materzynska, D.~Vazquez, and A.~M. Lopez, ``The synthia
  dataset: A large collection of synthetic images for semantic segmentation of
  urban scenes,'' in \emph{Proceedings of the IEEE conference on computer
  vision and pattern recognition}, 2016, pp. 3234--3243.

\end{thebibliography}

\end{document}